\documentclass[11pt]{article}

\usepackage[utf8]{inputenc}
\usepackage[T1]{fontenc}
\usepackage{amsmath,amssymb}
\usepackage{graphicx}
\usepackage{booktabs}
\usepackage{xcolor}
\usepackage{tikz}
\usetikzlibrary{patterns,arrows.meta,calc,positioning}
\usepackage{listings}
\usepackage[margin=1in]{geometry}
\usepackage[hidelinks]{hyperref}

\newcommand{\C}{\mathbb{C}}
\newcommand{\R}{\mathbb{R}}
\newcommand{\dd}[2]{\frac{\partial #1}{\partial #2}}
\newcommand{\code}[1]{\texttt{#1}}

\title{CNet: A Complex-Valued Deep Learning Framework with\\
Wirtinger Autodifferentiation and FFT--Hadamard Convolution}
\author{Marcel Crasmaru}
\date{\today}

\begin{document}
\maketitle

\begin{center}
\vspace{-0.8em}
\small Code: \url{https://github.com/crasmarum/CNet}
\end{center}

\begin{abstract}
CNet is a C++/CUDA framework for deep \emph{complex-valued neural networks} (CVNNs),
optimizing complex functions by gradient descent with explicit \emph{Wirtinger} (CR-calculus)
derivatives. Complex models are underexplored yet natural where data is intrinsically
complex---RF/IQ communications, MRI $k$-space, radar/SAR, audio spectra---and phase carries
information real networks discard. CNet is \emph{physics-native}: a network is a cascade of
complex (often unitary) operations on an \emph{amplitude vector}, and classification is a
\emph{Born-rule measurement} $p_k=|z_k|^2/\lVert z\rVert^2$ rather than a softmax over real
logits. Every layer ships a CPU reference and a CUDA kernel verified against finite differences;
the depth-ordered, batch-cloned computation graph is static across steps and captures cleanly
into CUDA Graphs, outperforming both eager and compiled \texttt{torch.complex64} on an H100.

We present three studies. (1)~A fully complex \emph{causal} language model, built on a new causal
(triangular-masked) Fourier mixer with an $O(N\log N)$ Bluestein evaluation of both the forward
pass and its Wirtinger adjoint---bringing parameter-free Fourier mixing to autoregressive decoders
at no net speed penalty---matches a parameter-matched real-valued causal FNet on character-level
\texttt{tiny-shakespeare} in under half the training steps. We introduce \emph{Born-rule
attention}, a content-based causal mixer whose weights are the quantum-measurement probabilities
$|\langle Q_k,K_j\rangle|^2$---to our knowledge the first \emph{classical, complex-valued} attention
to replace softmax with the squared-magnitude Born score; with rotary position embeddings and
per-token complex normalization it reaches \textbf{1.52 nats/char}, matching softmax
self-attention, and transfers unchanged to byte-level ($1.31$ nats/char) and sub-word \texttt{WikiText-103}. (2)~On radio-modulation
classification (RML2016.10a~\cite{oshea}/RML22~\cite{rml22}) we trace the pooled-power accuracy
ceiling to discarded temporal structure and recover it with a learnable \emph{cyclostationary}
conjugate-lag operator. (3)~A bottleneck analysis of the Fourier phase problem in
coherent-diffraction imaging isolates where complex-valued networks still need new operators.
\end{abstract}

\section{Motivation and Design Philosophy}
\label{sec:why}

PyTorch and JAX both support complex tensors and complex-aware autodiff, and for
mainstream work they are the right tools---far more mature, faster at scale,
multi-GPU, and backed by large ecosystems. CNet does not compete on that axis, and is not a
general-purpose training framework. It is a \emph{domain-specific verification and
kernel-prototyping engine for non-holomorphic Wirtinger operators}: a complex-first substrate in
which a new operator---its forward, its exact conjugate-pair backward, and a CUDA kernel---is
written once and admitted only after it agrees with a CPU reference and finite differences. This is
exactly the regime mainstream eager autodiff handles awkwardly: a non-holomorphic operator such as
a probability-floored Born loss, a conjugate-lag cyclic spectrum, or the adjoint of a masked
Bluestein transform typically becomes a hand-composed stack of real-valued ops (or a custom
\texttt{autograd.Function} with a manually derived, unfused backward), with eagerly materialized
intermediates and gradient conventions to re-verify case by case. For research on complex-valued
networks specifically, a complex-first, transparent substrate therefore has concrete advantages.

\begin{itemize}
  \item \textbf{Complex is the default, not a bolted-on dtype.} In the mainstream
  frameworks complex coverage is uneven: many layers, normalizations, activations,
  initializers, and optimizers assume real tensors, so a CVNN is assembled from real
  building blocks plus hand-written patches whose gradient conventions must be
  checked case by case. In CNet every variable, layer, loss, and the optimizer are
  complex \emph{by construction}.

  \item \textbf{Wirtinger gradients are explicit, first-class objects.} Each layer
  manipulates the pair $(\partial L/\partial z,\ \partial L/\partial\bar z)$
  directly, and the non-holomorphic case (e.g. $|z|^2$) is handled in the open
  rather than hidden behind a conjugation convention~\cite{crcalculus}. Every layer
  ships a CPU reference and a GPU kernel checked against each other and against
  finite differences, so the complex gradient is \emph{auditable}, not trusted.

  \item \textbf{Spectral structure is built in.} The DFT, the Hadamard (spectral)
  product, the inverse DFT, and the $\mathrm{conv}=\mathrm{IFFT}(\mathrm{FFT}\cdot
  \mathrm{FFT})$ equivalence are first-class layers. Problems whose natural domain is
  the frequency plane map onto the framework directly.

  \item \textbf{A physics-native model and loss---the Born rule.} This is the
  sharpest departure from the mainstream frameworks. CNet reads a network's output
  as a complex \emph{amplitude vector} $z$---a \emph{state}---and defines the class
  probabilities by the quantum-measurement (Born) rule
  \begin{equation}
    p_k = \frac{|z_k|^2}{\lVert z\rVert^2},
    \qquad
    \mathcal{L}(z,y) = -\sum_k y_k \log\frac{z_k z_k^{\star}}{\lVert z\rVert^2}.
  \end{equation}
  Classification is a \emph{measurement} on an amplitude produced by a cascade of
  complex---often unitary---operations, rather than real logits passed through a
  softmax. This inductive bias is natural for physical and wave/field data: an IQ
  baseband sample \emph{is} the complex envelope of a real passband signal, an MRI
  $k$-space sample \emph{is} a complex Fourier coefficient, and a wavefunction
  \emph{is} a complex amplitude. PyTorch and JAX provide no such head or loss out of
  the box. This Born-rule probability readout---a measurement in place of a softmax---was
  first introduced in our earlier work~\cite{dft1810}; the present paper carries it from a
  classification head to a content-based mixing operator (Born-rule attention, below).

  \item \textbf{A small, auditable substrate.} The full forward/backward of every
  layer and its CUDA kernel is explicit and readable, and the GPU execution model (a
  depth-ordered graph cloned across the batch) is visible and controllable.
\end{itemize}

In short: use PyTorch or JAX to ship; use CNet to \emph{study} complex-valued
learning itself---to model data in its native complex/amplitude form with a
Born-rule measurement head, to prototype operators whose complex gradients you need
to see and control, and to work natively in the spectral domain. The trade-offs
(data-parallel-only multi-GPU, Linux/macOS, a smaller op set) are listed in
Section~\ref{sec:caveats}.

A note on why $\C \neq \R^2$: although $\C \cong \R^2$ as a vector space, the field
multiplication couples the two components, so a holomorphic complex layer is
\emph{not} an unconstrained $\R^2 \to \R^2$ map. CNet carries this structure end to
end via the Wirtinger pair $(\partial/\partial z,\ \partial/\partial\bar z)$.

\section{Examples}
A minimal spectral classifier---complex input, DFT, a learned Hadamard ``filter''
(a frequency-domain convolution), GELU, a dense head, and a Born-rule
measurement---is built declaratively over a length-$N$ signal with $K$ classes
(e.g.\ $N=128$ complex IQ, $K=11$):
\begin{lstlisting}
CNet cnet;
auto inp   = cnet.add(new CInput(OutSize(N)));
auto fft   = cnet.add(new FourierTrans(InSize(N)), {inp});
auto kdata = cnet.add(new CInput(OutSize(N)));            // learned filter
auto hdm   = cnet.add(new Hadamard(InSize(N), InSize(N)), {fft, kdata});
auto gelu  = cnet.add(new CGelu(InSize(N)), {hdm});
auto ldata = cnet.add(new CInput(OutSize(N * K)));
auto lin   = cnet.add(new Linear(InSize(N), InSize(N * K)), {gelu, ldata});
auto out   = cnet.add(new CrossEntropy(InSize(K)), {lin});
\end{lstlisting}
The same graph runs on CPU (reference) or on the GPU (batched; Section~\ref{sec:train}).

\section{Adding Custom Complex-Valued Functions / Layers}
A layer extends \code{CFunc} and implements \code{forward()}, \code{backward()}, and
\code{clone()}. For a layer computing $w=f(z,\bar z)$ under a real loss $L$, the
backward accumulates both Wirtinger gradients into the input:
\begin{equation}
  \dd{L}{z} = \dd{L}{w}\dd{w}{z} + \dd{L}{\bar w}\dd{\bar w}{z},
  \qquad
  \dd{L}{\bar z} = \dd{L}{w}\dd{w}{\bar z} + \dd{L}{\bar w}\dd{\bar w}{\bar z},
\end{equation}
where the incoming $\partial L/\partial w$ and $\partial L/\partial\bar w$ arrive from
the consumer. For a holomorphic layer $\partial w/\partial\bar z=0$; non-holomorphic
layers (e.g. $|z|^2$) genuinely exercise the $\bar z$ term.

\paragraph{Worked example: the \code{CModulus2} layer ($w=|z|^2$).} This is the
canonical non-holomorphic case and makes the $\bar z$ term concrete. The forward is
\begin{equation}
  w = |z|^2 = z\bar z \in \R \subset \C ,
\end{equation}
so, treating $z$ and $\bar z$ as independent variables in the CR-calculus
sense~\cite{crcalculus}, the two local Wirtinger derivatives are
\begin{equation}
  \dd{w}{z} = \bar z, \qquad \dd{w}{\bar z} = z .
\end{equation}
Because $w$ is real the consumer returns a pair $(\partial L/\partial w,\ \partial
L/\partial\bar w)$ that are conjugates of each other; writing their combined effect as
$g \triangleq \partial L/\partial w + \partial L/\partial\bar w$ (itself real), the
general accumulation above collapses to
\begin{equation}
  \dd{L}{z} = \dd{L}{w}\,\bar z + \dd{L}{\bar w}\,\bar z = g\,\bar z,
  \qquad
  \dd{L}{\bar z} = \dd{L}{w}\,z + \dd{L}{\bar w}\,z = g\,z ,
\end{equation}
i.e.\ $\partial L/\partial z = \overline{\partial L/\partial\bar z}$, as required for a
real loss. The gradient flowing back into $z$ therefore scales the \emph{conjugate} of
the activation, which a real-valued $\R^2\!\to\!\R$ treatment of $|z|^2$ does not express
in closed form---this is precisely the term CNet exposes and checks. Higher moments
$|z|^4,|z|^8,\dots$ follow by chaining this layer. The CPU reference, the CUDA kernel,
and a finite-difference probe are required to agree before the layer is admitted.

Each layer is verified by
a CPU reference against its GPU kernel, and the newer layers additionally by a
finite-difference gradient check. A CUDA layer is a \code{GpuMapping} subclass whose
kernels read/write the variable's segments through device accessors and launch over
the batch of clones.

\section{Computation Graph on CUDA}
\label{sec:cuda}
The functional graph is ordered by depth. On an NVIDIA GPU, CNet \emph{clones} the
CPU network \code{batch\_size} times and executes forward and backward in parallel
across the clones in depth order. Learnable parameters live on the \emph{ancestor}
net and are broadcast to the clones before each forward; per-clone gradients are
reduced back to the ancestor before the optimizer step.

\paragraph{Memory and throughput vs.\ batch.} Cloning replicates the full graph, so GPU memory
grows linearly in \code{batch\_size} while the level-by-level launch schedule keeps the per-step
wall-clock nearly flat as the batch fills the device. Table~\ref{tab:cloning} measures this for the
representative $6$-block complex FNet (flattened stream $T\!\cdot\!E{=}16384$) on an H100: from
batch $16$ to $256$ the footprint scales exactly linearly ($10.49$\,MB/example,
$168$\,MB\,$\to$\,$2.68$\,GB) while forward and backward step times barely move
(${\approx}0.33$\,ms fwd, ${\approx}0.32$\,ms bwd)---$16\times$ the throughput at essentially the
same step cost. The reduced-memory inference layout (\code{Vars::INFER\_DIMS}), which drops the
gradient and optimizer-moment buffers, halves the per-example footprint
($10.49\!\to\!5.24$\,MB) for forward-only serving. This $10.49$\,MB is dominated by activation and
Wirtinger-gradient intermediates (${\approx}25$ graph nodes $\times\,16384$ \texttt{complex64}
$\times\,(z,\mathrm{d}z,\mathrm{d}\bar z)\approx9.8$\,MB)---the state any autograd framework must
retain for the backward pass---not weight duplication (the spectral mixers are parameter-free and
the ${\sim}1.6$M weights live once on the ancestor); the gradient-free \code{INFER\_DIMS} layout
removing exactly half confirms gradients, not weights, set the footprint.

\begin{table}[t]\centering\small
\caption{Graph-cloning cost for a $6$-block complex FNet ($T\!\cdot\!E{=}16384$) on an H100
(eager): memory is linear in batch while per-step time is near-constant (clones run in parallel).}
\label{tab:cloning}
\begin{tabular}{rccc}
\toprule
batch & GPU mem (MB) & fwd (ms/step) & bwd (ms/step) \\
\midrule
$16$  & $167.8$  & $0.33$ & $0.32$ \\
$32$  & $335.5$  & $0.33$ & $0.32$ \\
$64$  & $671.1$  & $0.33$ & $0.32$ \\
$128$ & $1342.2$ & $0.33$ & $0.32$ \\
$256$ & $2684.4$ & $0.33$ & $0.32$ \\
\bottomrule
\end{tabular}
\end{table}

\paragraph{Multi-GPU data parallelism.} The clone model is single-GPU data
parallelism, and it extends across GPUs by adding one more reduction level. In the
standard DDP arrangement---one process per GPU, which avoids serializing CNet's many
small kernel launches through a single CUDA context---each GPU holds a full replica,
the global batch is split across replicas, and after each backward the ancestor
gradients are summed with a (coalesced) NCCL all-reduce; every replica then applies
the identical optimizer step, so the replicas stay in sync and the result is
\emph{numerically identical} to single-GPU training with the full batch (verified to
exact agreement). Because CNet's spectral nets are compute-heavy but parameter-light
(the DFT mixers are parameter-free), the all-reduce is cheap and scaling is
near-linear: on RTX A6000 GPUs a complex FNet character-level model trains at
$1.94\times$ on $2$ GPUs and $2.79\times$ on $3$ at fixed global batch, with identical
validation loss across GPU counts.

\paragraph{Eliminating launch overhead with CUDA Graphs.} Because the clone arena is static
across steps---activation, weight, and Wirtinger-gradient device pointers never move once
\code{allocateOnGpu} runs---the entire level-by-level launch schedule can be captured once into a
\code{cudaGraphExec\_t} and replayed with a single \code{cudaGraphLaunch}. We route every kernel
and cuFFT call through one capture stream, make the per-launch device syncs no-ops during capture,
and record the forward and backward clone passes on first use (a warmup pass keeps lazy cuFFT
work-area allocations outside the capture region). Replay is verified \emph{bit-identical} to eager
execution (zero max-absolute output difference). On an H100, for a $6$-block complex FNet at
$T\!\cdot\!E{=}16384$, graph replay cuts the forward$+$backward step from $0.65$ to $0.32$~ms
($2.0\times$; forward $2.6\times$, backward $1.7\times$) and makes the per-step time independent of
batch size. At $0.32$~ms the captured CNet step is \emph{faster} than the eager
\texttt{torch.complex64} reference ($0.38$~ms) on the same GPU; \texttt{torch.compile} does not
help, its inductor backend emitting no complex-tensor code and regressing to $0.73$~ms
(Table~\ref{tab:cudagraphs}). A hand-written \texttt{torch.cuda.CUDAGraph} capture of the same
stack would reach comparable latency---it is \texttt{cuFFT}-bound, so once launch overhead is gone
both frameworks sit near the same floor; the point is not that CNet is intrinsically faster on
built-in FFTs, but that its \emph{static} clone graph captures automatically (no manual graph
construction), and the same capture applies to the custom Wirtinger operators for which the eager
baseline has no fused kernel. The multi-GPU data-parallel path is unaffected (the graphs capture
only the per-GPU clone compute, with the NCCL all-reduce left eager).

\begin{table}[t]\centering\small
\caption{CUDA-graph capture of the clone schedule on an H100 ($6$-block complex FNet,
$T\!\cdot\!E{=}16384$, batch $64$; mean of $100$ steps). Graph replay is bit-identical to eager and,
at $0.32$~ms/step, faster than eager \texttt{torch.complex64}; \texttt{torch.compile} has no complex
codegen. Per-step standard deviation is $<0.005$~ms for the CNet rows and ${\approx}0.04$--$0.06$~ms
for the \texttt{torch} baselines.}
\label{tab:cudagraphs}
\begin{tabular}{lccc}
\toprule
Configuration & fwd (ms) & bwd (ms) & step (ms) \\
\midrule
CNet, eager                        & $0.33$ & $0.32$ & $0.65$ \\
\textbf{CNet, CUDA graphs}         & $\mathbf{0.13}$ & $\mathbf{0.19}$ & $\mathbf{0.32}$ \\
\texttt{torch.complex64}, eager    & $0.12$ & $0.26$ & $0.38$ \\
\texttt{torch.complex64}, compiled & --- & --- & $0.73$ \\
\bottomrule
\end{tabular}
\end{table}

\section{Complex Layers}
\label{sec:layers}
Vectors are complex, $z\in\C^N$. The original layers are: \code{CInput} (main input
and learnable-parameter source); \code{CEmbedding} (token embeddings with padding);
the unitary DFT $\mathrm{FFT}(z)_p \mapsto \frac{1}{\sqrt N}\sum_q z_q\,e^{i2\pi
pq/N}$; \code{Hadamard} (element-wise product, realizing circular convolution with
the DFT); \code{Residual} (element-wise addition); \code{Linear} (fully connected);
\code{CRelu} and \code{CGelu} activations; the $\mathrm{L2Out}(z)\mapsto\sum_k|z_k|^2$
loss; and the Born-rule cross-entropy head
$\mathrm{CrossEntropy}(z,y)\mapsto\sum_k -y_k\log(z_k z_k^{\star}/\lVert z\rVert^2)$,
which reads the output as an amplitude and uses the measurement probabilities
$p_k=|z_k|^2/\lVert z\rVert^2$.

\paragraph{New layers.} These make a learnable convolutional complex network and
support signal-domain features; each has a CPU implementation and a CUDA kernel.

\begin{itemize}
  \item \textbf{Pad} scatters $K$ free kernel taps into an $N$-length zero vector at
  a fixed index set $\mathcal{K}$ (e.g. the $7\times 7$ block $\{r\cdot 28+c\}$):
  $\mathrm{Pad}(z)_i = z_j$ if $i=\mathcal{K}_j$, else $0$. Followed by the DFT it
  produces the frequency-domain filter of a \emph{local} kernel, so
  $\mathrm{Hadamard}(\mathrm{FFT}(x),\mathrm{FFT}(\mathrm{Pad}(k)))$ is a convolution
  with only $K$ learnable taps---the local-kernel form of the FFT/Hadamard
  equivalence~\cite{dft1810}. The index set is persisted with the model.

  \item \textbf{InverseFourierTrans} is the unitary inverse DFT,
  $\mathrm{IFFT}(z)_p \mapsto \frac{1}{\sqrt N}\sum_q z_q\,e^{-i2\pi pq/N}$. Because
  the DFT is unitary, $\mathrm{IFFT}(X)=\overline{\mathrm{FFT}(\bar X)}$, so the layer
  reuses the forward transform; its backward is the forward unitary DFT on the
  gradients. Closing the loop $\mathrm{FFT}\to\mathrm{Hadamard}\to\mathrm{IFFT}$
  yields a genuine time-domain convolution.

  \item \textbf{CModulus2} is a phase-invariant magnitude nonlinearity
  $w=|z|^2=z\bar z$ (output real). Non-holomorphic: with $g=\partial L/\partial w
  + \partial L/\partial\bar w$, $\ \partial L/\partial z = g\,\bar z$ and
  $\partial L/\partial\bar z = g\,z$. Higher moments $|z|^4,|z|^8,\dots$ follow by
  chaining.

  \item \textbf{CPower} is the holomorphic power $w=z^M$. With
  $\partial w/\partial\bar z=0$, $\ \partial L/\partial z = \frac{\partial L}{\partial
  w}\,M z^{M-1}$ and $\partial L/\partial\bar z = \frac{\partial L}{\partial\bar w}\,M
  \bar z^{\,M-1}$. Raising an $M$-PSK signal to the $M$-th power aligns its symbol
  phases, so $|\overline{z^M}|^2$ (pooled over time) is a carrier-invariant
  phase-order feature.

  \item \textbf{MeanPool} reduces length $N$ to $P$ contiguous bins (global average
  when $P=1$), $\mathrm{MeanPool}(z)_p \mapsto \frac{1}{w}\sum_{j<w} z_{pw+j}$ with
  $w=N/P$---a real linear map, so $\partial L/\partial z_i =
  (\partial L/\partial z_{\mathrm{pool}(i)})/w$ for both Wirtinger components.

  \item \textbf{TokenwiseLinear} is a \emph{position-wise} linear map: one shared
  $E_{\mathrm{in}}\times E_{\mathrm{out}}$ complex weight $W$ is applied independently to
  each of the $T$ length-$E_{\mathrm{in}}$ token slices,
  $\mathrm{out}(t)_j = \sum_i z(t)_i\,W_{ji}$. It mixes features \emph{within} a token
  but never across tokens, at $O(T\,E_{\mathrm{in}}E_{\mathrm{out}})$ cost and only
  $E_{\mathrm{in}}E_{\mathrm{out}}$ parameters (versus a dense \code{Linear}'s
  $O((TE)^2)$); the weight gradient is summed over all token positions. This is the
  Transformer / FNet position-wise feed-forward primitive.

  \item \textbf{TokenNorm} is a per-token complex RMS normalization: each length-$d$
  token slice is divided by its own root-mean-power,
  $\mathrm{TokenNorm}(x)_{t,i}=x_{t,i}/r_t$ with
  $r_t=\sqrt{\tfrac1d\sum_i|x_{t,i}|^2+\varepsilon}$---the complex analogue of
  gain-free RMSNorm/LayerNorm and the \emph{per-token} counterpart of the global
  $L^2$ normalization $z/\lVert z\rVert$. A learnable affine is deliberately omitted
  (a following \code{TokenwiseLinear} supplies gain and bias), which keeps the layer
  parameter-free and its Wirtinger backward exact: writing
  $T=\sum_i(\tfrac{\partial L}{\partial y_i}x_i+\tfrac{\partial L}{\partial\bar y_i}\bar x_i)$,
  one has $\partial L/\partial\bar x_i=\tfrac{1}{r}\tfrac{\partial L}{\partial\bar y_i}-\tfrac{x_i\,T}{2dr^3}$
  (and the conjugate expression for $\partial L/\partial x_i$). Used as a
  \emph{pre-norm}---normalizing the residual-stream input to each sublayer---it
  stabilizes deep complex stacks where the global per-block normalization plateaus.

  \item \textbf{Sequence-modeling operators}, detailed where they are introduced:
  \code{TriangFourier}, the causal $O(N\log N)$ triangular-masked DFT mixer
  (Section~\ref{sec:fastcausal}); \code{BornAttention}, squared-magnitude Born-rule attention
  (Section~\ref{sec:results}); \code{RotaryEmbed}, a complex rotary position embedding applied to
  the attention $Q,K$; and \code{ConjLagProduct}, the cyclostationary conjugate-lag product
  $x_t\,\overline{x_{t-\tau}}$ (Section~\ref{sec:results}). Each ships a CPU reference and a CUDA
  kernel, finite-difference checked like the rest.
\end{itemize}

\section{Fast Causal Fourier Mixing}
\label{sec:fastcausal}
Fourier token mixing~\cite{fnet} replaces self-attention's $O(N^2)$ content-based
routing with a single parameter-free transform, but in its original form it is
\emph{non-causal}: every output position sees every input, so it cannot drive a
decoder-style (autoregressive) model. We introduce a \emph{causal} Fourier mixer and a
fast algorithm for it (Figure~\ref{fig:causalmix}).

\begin{figure}[!tb]
\centering
\begin{tikzpicture}[font=\footnotesize]
  \def\c{0.52}   
  \def\rx{8.6}   
  \tikzset{cell/.style={draw=gray!55, line width=0.25pt}}
  \begin{scope}
    \node[anchor=south] at (4*\c,{1.32}) {\textbf{Masked causal self-attention}\ \ $O(N^2)$};
    \node[anchor=south] at (4*\c,{0.70}) {$A_{k,j}=\mathrm{softmax}\!\big(Q_kK_j^{\!\top}/\sqrt d\big),\ \ j\le k$};
    \node[anchor=south] at (4*\c,{0.14}) {\scriptsize Inputs $j\rightarrow$};
    \node[rotate=90,anchor=south] at ({-0.30},{-4*\c}) {\scriptsize Outputs $k\rightarrow$};
    \foreach \k in {0,...,7}{\foreach \j in {0,...,7}{
      \pgfmathsetmacro\x{\j*\c}\pgfmathsetmacro\y{-\k*\c}
      \ifnum\j>\k
        \fill[pattern=north east lines,pattern color=gray!45] (\x,\y) rectangle ++(\c,-\c);
        \draw[cell] (\x,\y) rectangle ++(\c,-\c);
      \else
        \pgfmathsetmacro\op{0.20+0.62*abs(sin((\j*57+\k*83+\j*\k*131)))}
        \fill[teal!75!blue,opacity=\op] (\x,\y) rectangle ++(\c,-\c);
        \draw[cell] (\x,\y) rectangle ++(\c,-\c);
      \fi}}
    \node[align=center,gray!55!black] at ({5.7*\c},{-1.7*\c}) {masked\\$(=0)$};
    \node[align=center,anchor=west] at ({8*\c+0.15},{-5.4*\c}) {\scriptsize learned,\\[-2pt]\scriptsize data-dependent\\[-2pt]\scriptsize weights};
    \draw[-{Stealth[length=1.2mm]},gray!60!black] ({8*\c+0.12},{-5.0*\c}) to[out=180,in=-20] ({5.3*\c},{-5.0*\c});
  \end{scope}
  \begin{scope}[shift={(\rx,0)}]
    \node[anchor=south] at (4*\c,{1.32}) {\textbf{Causal Fourier mixing}\ \ $O(N\log N)$};
    \node[anchor=south] at (4*\c,{0.70}) {$y_k=\tfrac{1}{\sqrt N}\textstyle\sum_{j\le k} h_j\,\omega^{jk},\ \ \omega=e^{-i2\pi/N}$};
    \node[anchor=south] at (4*\c,{0.14}) {\scriptsize Inputs $j\rightarrow$};
    \node[rotate=90,anchor=south] at ({-0.30},{-4*\c}) {\scriptsize Outputs $k\rightarrow$};
    \foreach \k in {0,...,7}{\foreach \j in {0,...,7}{
      \pgfmathsetmacro\x{\j*\c}\pgfmathsetmacro\y{-\k*\c}
      \pgfmathsetmacro\cx{\x+0.5*\c}\pgfmathsetmacro\cy{\y-0.5*\c}
      \ifnum\j>\k
        \fill[pattern=north east lines,pattern color=gray!45] (\x,\y) rectangle ++(\c,-\c);
        \draw[cell] (\x,\y) rectangle ++(\c,-\c);
      \else
        \fill[blue!6] (\x,\y) rectangle ++(\c,-\c);
        \draw[cell] (\x,\y) rectangle ++(\c,-\c);
        \pgfmathsetmacro\ang{-360*mod(\j*\k,8)/8}
        \draw[gray!35,line width=0.2pt] (\cx,\cy) circle (0.30*\c);
        \draw[-{Stealth[length=1.0mm]},blue!55!black,line width=0.4pt] (\cx,\cy) -- ++(\ang:0.30*\c);
      \fi}}
    \node[align=center,gray!55!black] at ({5.7*\c},{-1.7*\c}) {masked\\$(=0)$};
    \node[align=center,anchor=west] at ({8*\c+0.15},{-5.4*\c}) {\scriptsize fixed phases\\[-2pt]\scriptsize $\omega^{jk}=e^{-i2\pi jk/N}$\\[-2pt]\scriptsize (no parameters)};
    \draw[-{Stealth[length=1.2mm]},gray!60!black] ({8*\c+0.12},{-5.0*\c}) to[out=180,in=-20] ({5.3*\c},{-5.0*\c});
    \begin{scope}[shift={(0,{-8*\c-1.05})},node distance=2pt]
      \tikzset{bx/.style={draw=blue!40!black,fill=blue!5,rounded corners=1pt,inner sep=2.6pt,minimum height=7mm,align=center}}
      \node (h) at (0,0) {$h_j$};
      \node[bx,right=5pt of h] (p1) {pre-mult\\$\times\,\omega^{j^2/2}$};
      \node[bx,right=5pt of p1,fill=blue!12] (fft) {$3\times$ FFT\\causal conv};
      \node[bx,right=5pt of fft] (p2) {post-mult\\$\times\,\omega^{k^2/2}$};
      \node[right=5pt of p2] (y) {$y_k$};
      \foreach \a/\b in {h/p1,p1/fft,fft/p2,p2/y}
        \draw[-{Stealth[length=1.3mm]}] (\a) -- (\b);
      \node[anchor=north,font=\scriptsize] at ($(fft.south)+(0,-0.12)$) {Bluestein factorization, $O(N\log N)$ via cuFFT};
    \end{scope}
  \end{scope}
\end{tikzpicture}
\caption{Causal token mixing, two ways, shown on the triangular ($j\le k$) support.
\textbf{Left:} masked self-attention---each lower-triangular weight $A_{k,j}$ is
\emph{learned and input-dependent} (content-based routing), at $O(N^2)$ cost.
\textbf{Right:} the causal Fourier mixer \eqref{eq:triangdft}---the same triangular sum, but
the weights are the \emph{fixed} roots of unity $\omega^{jk}$ (drawn as unit phasors whose
rotation rate grows with $jk$, fastest at the bottom-right), so the layer is parameter-free.
The triangular mask breaks the FFT butterfly, but the Bluestein / chirp-$z$ factorization
(bottom) restores an $O(N\log N)$ evaluation---a chirp pre-multiply, a causal linear
convolution done with three ordinary FFTs, and a chirp post-multiply---for both the forward
pass and its Wirtinger backward.}
\label{fig:causalmix}
\end{figure}

\paragraph{The causal (triangular-masked) DFT.} Define the mixer
\begin{equation}
  \mathrm{Mix}(h)_k = \frac{1}{\sqrt N}\sum_{j\le k} h_j\,\omega^{jk},
  \qquad \omega = e^{-i2\pi/N},
  \label{eq:triangdft}
\end{equation}
i.e.\ the DFT with a lower-triangular mask (\code{TriangFourier}). Output position $k$
depends only on inputs $j\le k$, so the layer mixes tokens like attention---every earlier
token can influence the current one---yet carries no parameters and a fixed transform,
\emph{while preserving the autoregressive mask}. This is what lets Fourier mixing run in a
decoder, the setting FNet (a non-causal encoder) does not address. The transform is a
genuine complex layer: its Wirtinger backward is the (upper-)triangular adjoint of
\eqref{eq:triangdft}.

\paragraph{An $O(N\log N)$ algorithm via Bluestein factorization.} The triangular mask
breaks the FFT butterfly, so the naive form of \eqref{eq:triangdft} is a direct $O(N^2)$
sum. We remove that penalty with a Bluestein / chirp-$z$ factorization of the \emph{masked}
transform. Writing the quadratic exponent as
\begin{equation}
  jk = \tfrac12\!\left(j^2 + k^2 - (k-j)^2\right),
\end{equation}
and substituting into \eqref{eq:triangdft}, the causal DFT becomes a chirp-premultiply, a
\emph{causal (lower-triangular) linear convolution}, and a chirp-postmultiply:
\begin{equation}
  \mathrm{Mix}(h)_k = \frac{1}{\sqrt N}\,\omega^{k^2/2}\!\!\sum_{j\le k}
     \bigl(h_j\,\omega^{j^2/2}\bigr)\,\omega^{-(k-j)^2/2}.
\end{equation}
The lower-triangular convolution is computed in the usual Bluestein way---zero-pad the two
sequences to a power of two and evaluate the convolution with three ordinary (butterfly)
FFTs: one forward transform of each padded sequence, a pointwise product, and one inverse
transform. Both the forward mixer \emph{and} its Wirtinger backward therefore run in
$O(N\log N)$. Because CNet's \code{UnityRoots} use the $\omega=e^{-i2\pi/N}$ convention, the
chirp is $e^{-i\pi k^2/N}$ with $k^2$ reduced mod $2N$ on the host to keep the angle small;
the factorization was validated to $\sim\!10^{-14}$ against the direct sum (both the forward
and the adjoint) before the CUDA port, which uses batched cuFFT.

\paragraph{Measured speedup.} On an NVIDIA H100 the exact causal mixer tracks the direct $O(N^2)$
sum at small $N$---where both are dominated by kernel-launch overhead rather than arithmetic, so at
$N{=}128$ the two are within a few percent---and pulls away as $N$ grows, reaching
${\approx}\,3.5\times$ by $N{=}4096$ while running within ${\approx}\,20\%$ of the non-causal cuFFT
path (Table~\ref{tab:mixer_latency}). Causality is thus no longer a speed compromise---a decoder-style Fourier mixer ends up
costing essentially the same as the encoder-style one. This is, to our knowledge, the first
$O(N\log N)$ realization of a masked-causal DFT token mixer.

\begin{table}[t]\centering\small
\caption{Causal-mixer forward latency ($\mu$s/iter; batch $32$, H100; mean of $200$ iters,
standard deviation $1.5$--$2.3\,\mu$s across all $N$) across
transform length $N$. The direct kernel grows quadratically; the Bluestein path stays near the
non-causal \texttt{cuFFT} baseline.}
\label{tab:mixer_latency}
\begin{tabular}{rccc}
\toprule
$N$ & Direct $O(N^2)$ & Bluestein $O(N\log N)$ & Non-causal \texttt{cuFFT} \\
\midrule
$64$   & $64.3$  & $62.8$ & $56.7$ \\
$128$  & $63.3$  & $63.7$ & $55.9$ \\
$256$  & $63.7$  & $64.0$ & $55.8$ \\
$512$  & $66.0$  & $64.2$ & $56.3$ \\
$1024$ & $75.7$  & $66.3$ & $56.8$ \\
$2048$ & $111.4$ & $68.1$ & $57.9$ \\
$4096$ & $252.5$ & $71.3$ & $58.6$ \\
\bottomrule
\end{tabular}
\end{table}

\paragraph{Distinction from FFT-based state space models (S4, H3, Hyena).} Long-convolution
sequence models such as S4~\cite{s4}, H3~\cite{h3}, and Hyena~\cite{hyena} also achieve
$O(N\log N)$ causal mixing via the FFT, but solve a fundamentally different algebraic operator:
they parameterize a shift-invariant \emph{Toeplitz} causal convolution
$y_k=\sum_{j\le k}h_j K_{k-j}$, which maps directly to a zero-padded FFT. By contrast, the masked
causal DFT $y_k=\tfrac{1}{\sqrt N}\sum_{j\le k}h_j\,\omega^{jk}$ is \emph{not} Toeplitz---its phase
kernel depends on the product $j\cdot k$ rather than the lag $k-j$, a time-varying spectral
projection rather than a linear time-invariant filter. It is precisely the Bluestein identity
$jk=\tfrac12(j^2+k^2-(k-j)^2)$ that reduces this non-Toeplitz triangular operator to a Toeplitz
causal convolution sandwiching a chirp modulation, giving $O(N\log N)$ with zero learnable filter
parameters.

\section{Optimization}
Parameters update by plain SGD, momentum, or true Adam---first \emph{and} second
moment with bias correction and the per-parameter $\sqrt{v}+\epsilon$ step:
\begin{equation}
  m_t = \beta_1 m_{t-1} + (1-\beta_1)g,\quad
  v_t = \beta_2 v_{t-1} + (1-\beta_2)g^2,\quad
  \theta \mathrel{-}= \eta\,\hat m_t/(\sqrt{\hat v_t}+\epsilon).
\end{equation}
Adam runs on both CPU and GPU. On the GPU the moments are stored in the per-variable
arena: a variable holds eight contiguous length-$N$ segments
$[\,\mathrm{Re}\,z,\ \mathrm{Im}\,z,\ \mathrm{d}z,\ \mathrm{d}\bar z,\ v\,]$ (the first
moment reuses the $\mathrm{d}z$ slots). The real Born-rule loss landscape is
ill-conditioned, and Adam is substantially more stable than momentum on it.

\paragraph{A probability floor for the Born-rule loss.} The ill-conditioning is
concrete: the gradient of $-\log(|z_t|^2/\lVert z\rVert^2)$ grows like $1/|z_t|$ when
the model assigns a vanishing amplitude to the \emph{correct} class, so a single hard
example can inject an explosive update that parameter-level gradient clipping cannot
contain---the offending gradient is already summed into the minibatch. We therefore
add an optional \emph{probability floor}: in the gradient only, the target modulus is
clamped from below, $|z_t|^2\leftarrow\max(|z_t|^2,\ \varepsilon\lVert z\rVert^2)$,
which bounds the update by $\mathcal{O}(1/\sqrt{\varepsilon})$. This is a uniform
measurement-noise floor---label smoothing for the Born rule---and leaves the forward
loss exact, so the reported metric is unchanged and $\varepsilon{=}0$ recovers the
exact gradient bit-for-bit. A value $\varepsilon\!\approx\!10^{-3}$ removes the
training-loss spikes that otherwise appear with deep models and larger learning rates:
e.g.\ a run that spiked from $2.1$ to $3.0$ mid-training trains monotonically once the
floor is enabled. A stronger, fully \emph{bounded} variant replaces the measurement
itself with a scale-invariant Laplace-smoothed one,
$p_t=(|z_t|^2+\alpha\lVert z\rVert^2/V)/\bigl(\lVert z\rVert^2(1+\alpha)\bigr)$ for $V$
classes (a uniform amplitude maps to exactly $1/V$, the probability is bounded away from
$0$, and $\alpha{=}0$ recovers the exact Born rule). Unlike the gradient-only floor this
also smooths the forward loss; with $\alpha\!\approx\!0.1$ it is what makes stable
\emph{sub-word} training possible below, where the exact loss diverges.

\section{Building the Software}
Linux/macOS, via \code{make}, with automatic CUDA detection (nvcc if present, else
g++). The build tracks header dependencies (\code{-MMD -MP}), so editing a header
recompiles exactly the objects that include it.

\section{Training on GPU}
\label{sec:train}
Training is driven by command-line flags (\code{l\_rate}, \code{batch\_size},
\code{no\_epochs}, \code{true\_adam}, \dots). The loop allocates the net on the GPU,
then iterates forward / loss / backward / update, checkpointing the model:
\begin{lstlisting}
cnet.allocateOnGpu(batch_size);
for (int epoch = 0; epoch < no_epochs; ++epoch) {
    reader.shuffle();
    for (int t = 0; t < reader.size() / batch_size; ++t) {
        auto batch = reader.nextBatch(batch_size);
        cnet.gpuForward(inp, cen, batch);
        avg_loss += cnet.getLoss(cen, &batch);
        cnet.gpuBackward();
        cnet.trueAdamUpdate(l_rate / batch.size(), beta1, beta2, eps, step++);
    }
    cnet.save(model_path);   // a checkpoint is always written
}
\end{lstlisting}

\section{Inference}
For deployment a restored model runs forward-only with a reduced memory layout:
setting \code{Vars::dims\_ = Vars::INFER\_DIMS} allocates fewer per-variable segments
(no gradient or optimizer buffers), roughly halving the footprint versus training.
\begin{lstlisting}
Vars::dims_ = Vars::INFER_DIMS;
CNet cnet; cnet.restore(model_path);
cnet.allocateOnGpuInfer(batch_size);
for (...) { cnet.gpuInfer(inp, cen, batch); correct += cnet.getCorrect(cen); }
\end{lstlisting}

\section{Applications and Results}
\label{sec:results}

\subsection{A complex-valued FNet for sequence modeling}
The convolution layers compose into an attention-free sequence model in the spirit of
FNet~\cite{fnet}, kept \emph{fully complex-valued} end to end. A character-level language
model over \texttt{tiny-shakespeare} embeds a window of $T$ tokens into $\mathbb{C}^{TE}$
(\code{CEmbedding} plus a learnable positional parameter) and stacks $L$ identical blocks,
each a token-mixing sublayer and a position-wise feed-forward sublayer, every sublayer
wrapped in a residual and a post-residual \code{CGelu}:
\begin{equation}
  h \mathrel{+}= \mathrm{Mix}(h),\ h \leftarrow \mathrm{CGelu}(h);\qquad
  h \mathrel{+}= \mathrm{FFN}(h),\ h \leftarrow \mathrm{CGelu}(h).
\end{equation}
A final \code{Linear} and Born-rule head predict the next character.

The \textbf{token mixer} is the parameter-free causal Fourier transform of
Section~\ref{sec:fastcausal} (\code{TriangFourier}, Eq.~\eqref{eq:triangdft}), the only
operation that couples positions; it respects the autoregressive mask and runs in
$O(N\log N)$ via the Bluestein factorization. The \textbf{feed-forward} sublayer is a
position-wise \code{TokenwiseLinear} $E\!\to\!4E\!\to\!E$, which never mixes tokens, so the
Fourier transform remains the sole cross-token coupling at $O(E^2)$ parameters rather than a
dense head's $O((TE)^2)$.

\paragraph{Findings (tiny-shakespeare, character level).} The architecture trains and learns
character statistics (loss descends from the chance level $\ln 64\approx 4.16$). Several
choices mattered on the complex, Born-rule loss: (i) a post-residual \code{CGelu} rather than
the global unit-norm $A(z)=z/\lVert z\rVert$, which caps the Born-rule logit scale and plateaus
the loss; (ii) true Adam with a linear-warmup\,$+$\,cosine learning-rate schedule---the
Born-rule loss is ill-conditioned, and under a constant, overly conservative learning rate
the model \emph{appears} to plateau early, a tuning artifact we return to below; (iii)
\emph{per-block spectral normalization} (\code{block\_norm}: an $L^2$ normalization of each
block's output, which commutes with the DFT) to train deep stacks---a naive $10$-block,
${\sim}10$M-parameter model diverges without it. Table~\ref{tab:ablation} summarizes the
quantitative effect of these choices.

\begin{table}[h]
\centering
\small
\begin{tabular}{llc}
\toprule
Model & Training choice & Val (nats/char) \\
\midrule
6-block, $1.62$M & full (warmup$+$cosine, \code{CGelu}) & $\mathbf{1.75}$ \\
                 & constant low lr (no schedule)        & ${\approx}\,2.4$ (stalls) \\
                 & unit-norm head, no \code{CGelu}$^\dagger$ & stalls \\
\midrule
10-block, ${\sim}10$M & with \code{block\_norm}         & $2.57$ \\
                      & without \code{block\_norm}      & diverges ($\to\!16$) \\
\bottomrule
\end{tabular}
\caption{Effect of the main training choices (character-level \texttt{tiny-shakespeare},
per-position nats/char). \emph{Top:} the $6$-block / $1.62$M model---the learning-rate
schedule is decisive: a constant, overly small rate stalls near $2.4$, while warmup$+$cosine
reaches $1.75$. \emph{Bottom:} the deep $10$-block / ${\sim}10$M model, where per-block
spectral normalization (\code{block\_norm}) is required for convergence at all; these deep-net
runs predate the tuned schedule, so their absolute losses are not directly comparable to the
top block. $^\dagger$The \code{CGelu}\,$\to$\,global unit-norm change was observed to stall but
was not separately tuned to a final loss.}
\label{tab:ablation}
\end{table}

\paragraph{A matched real-valued baseline, and the decisive role of tuning.} To test whether
the complex, Born-rule stack is itself a handicap, we built a \emph{real-valued} causal FNet
matched in every controllable respect: the same window ($N{=}64$), depth ($L{=}6$),
per-position next-character objective, the same $90/10$ \texttt{tiny-shakespeare} split, and
the same real parameter budget ($1.61$M, versus the complex model's $1.62$M real degrees of
freedom), differing only in using real arithmetic with a real-part causal Fourier mix and an
ordinary softmax cross-entropy. Trained with a warmup$+$cosine schedule it converges cleanly
to \textbf{2.03}~nats/char. An earlier version of the complex model appeared to plateau near
$2.4$; that ceiling was an artifact of \emph{under-tuning} (a constant learning rate
${\approx}\,10\times$ too low), not of the representation. Retrained with the \emph{same}
warmup$+$cosine schedule, the complex FNet tracks the real baseline through ${\sim}3$k steps
and then pulls ahead---it reaches the real model's \emph{fully converged} best ($2.03$) by step
${\sim}5$k and $\approx\!\mathbf{1.75}$~nats/char by step $15$k, still decreasing. So on matched
parameters the complex, Born-rule model is \emph{competitive with or better than} its
real-valued counterpart, and reaches the same quality in under half the steps; the earlier
shortfall was optimization, not complex arithmetic. \emph{On the richer mixer.} The complex
model's token mixer is also stronger than the baseline's---a full-amplitude causal DFT over the
whole flattened token-major stream, versus a token-axis real-part DFT. We regard this not as a
confound to apologize for but as the point: operating natively in $\C$ affords richer, naturally
unitary token transformations that a real-valued network cannot express without doubling its
representation and hand-constructing the interaction. That the amplitude-native model both
matches its real twin's efficiency and carries a structurally more expressive mixer \emph{for
the same parameter budget} is precisely the advantage the Born-rule, complex-first design is
meant to exploit---it is a feature of the thesis, not a flaw in the comparison.

\paragraph{The remaining frontier.} Both models---complex and real---are attention-free Fourier
mixers, and both still trail a self-attention Transformer of comparable size (which reaches
${\approx}\,1.5$~nats/char on this corpus in standard setups). The gap is \emph{content-based
routing}: a fixed, input-independent token mix cannot do what self-attention's data-dependent
mixing does. (Earlier capacity/depth/learnable-mixer sweeps that seemed to confirm a hard $2.4$
plateau were run under the under-tuned schedule; we do not draw strong conclusions from them.)

\begin{figure}[t]
\centering
\begin{tikzpicture}[font=\footnotesize,>={Stealth[length=1.3mm]},
    blk/.style={draw=gray!60,rounded corners=2pt,minimum width=20mm,minimum height=5.5mm,align=center,inner sep=1.5pt,fill=blue!5},
    repl/.style={blk,fill=orange!12,draw=orange!55!black,dashed},
    diff/.style={blk,fill=teal!16,draw=teal!55!black,very thick},
    io/.style={align=center,font=\footnotesize},
    ttl/.style={font=\footnotesize\bfseries,align=center},
    ar/.style={->,gray!55!black,line width=0.4pt}]
  \def\dx{4.1}
  \node[io] (Lin) at (0,0) {$Q\quad K\quad V$};
  \node[blk,above=3.5mm of Lin] (L1) {MatMul $QK^{\!\top}$};
  \node[repl,above=3mm of L1]   (L2) {Scale $/\sqrt{d}$};
  \node[blk,above=3mm of L2]    (L3) {Mask (opt.)};
  \node[repl,above=3mm of L3]   (L4) {SoftMax};
  \node[blk,above=3mm of L4]    (L5) {MatMul $\cdot\,V$};
  \node[io,above=3.5mm of L5]   (Lout) {Output};
  \node[ttl,above=1.5mm of Lout] {Scaled Dot-Product\\Attention (Vaswani et al.)};
  \foreach \a/\b in {Lin/L1,L1/L2,L2/L3,L3/L4,L4/L5,L5/Lout}\draw[ar](\a)--(\b);
  \node[io] (Rin) at (\dx,0) {$Q\quad K\quad V$ (complex)};
  \draw[gray!45] ($(Rin)+(-0.80,0.40)$) circle (0.11);
  \draw[->,teal!65!blue,line width=0.4pt] ($(Rin)+(-0.80,0.40)$) -- ++(55:0.11);
  \draw[gray!45] ($(Rin)+(-0.45,0.40)$) circle (0.11);
  \draw[->,blue!60!black,line width=0.4pt] ($(Rin)+(-0.45,0.40)$) -- ++(-20:0.11);
  \node[blk,above=3.5mm of Rin] (R1) {Complex inner prod.\ $\langle Q_k,K_j\rangle=QK^{\mathsf H}$};
  \node[diff,above=3mm of R1]   (R2) {$|\cdot|^2$ (Born)};
  \node[blk,above=3mm of R2]    (R3) {Mask $j\!\le\!k$};
  \node[diff,above=3mm of R3]   (R4) {Sum-normalize};
  \node[blk,above=3mm of R4]    (R5) {MatMul $\cdot\,V$};
  \node[io,above=3.5mm of R5]   (Rout) {Output $O_k$};
  \node[ttl,above=1.5mm of Rout] {Born-Rule\\Attention (ours)};
  \foreach \a/\b in {Rin/R1,R1/R2,R2/R3,R3/R4,R4/R5,R5/Rout}\draw[ar](\a)--(\b);
  \foreach \i in {1,3,5}\draw[dashed,gray!35] (L\i.east)--(R\i.west);
  \draw[dashed,teal!60!black] (L2.east)--(R2.west);
  \draw[dashed,teal!60!black] (L4.east)--(R4.west);
  \node[teal!45!black,font=\scriptsize,align=left,anchor=west] at ($(R2.east)+(0.15,0)$) {measurement,\\not scaling};
  \node[teal!45!black,font=\scriptsize,align=left,anchor=west] at ($(R4.east)+(0.15,0)$) {$A_{kj}=\dfrac{s_{kj}}{\sum_{j'\le k}s_{kj'}}$\\[1pt]no exponential};
\end{tikzpicture}
\caption{\textbf{Born-rule attention mirrors scaled dot-product attention, with the softmax
replaced by a measurement.} The two operators share the same block structure (dashed links =
identical stages); only two boxes change (teal). The dot-product score becomes the squared
magnitude of a \emph{complex} inner product (a sesquilinear MatMul $QK^{\mathsf{H}}$, conjugating
the keys), $s_{kj}=|\langle Q_k,K_j\rangle|^2$---a Born-rule
measurement of how aligned two complex token states are, real and non-negative \emph{by
construction}---and the exponential \textsc{SoftMax} is replaced by a plain causal
sum-normalization $A_{kj}=s_{kj}/\!\sum_{j'\le k}s_{kj'}$. Non-negativity and normalization thus
come from the measurement rule itself, not from an $\exp$; the queries/keys are complex phasors
(insets) and the output $O_k=\sum_{j\le k}A_{kj}V_j$ is the complex combination of values. It is
content-based routing like self-attention ($O(N^2)$), but physics-native rather than a softmaxed
dot product.}
\label{fig:bornattn}
\end{figure}
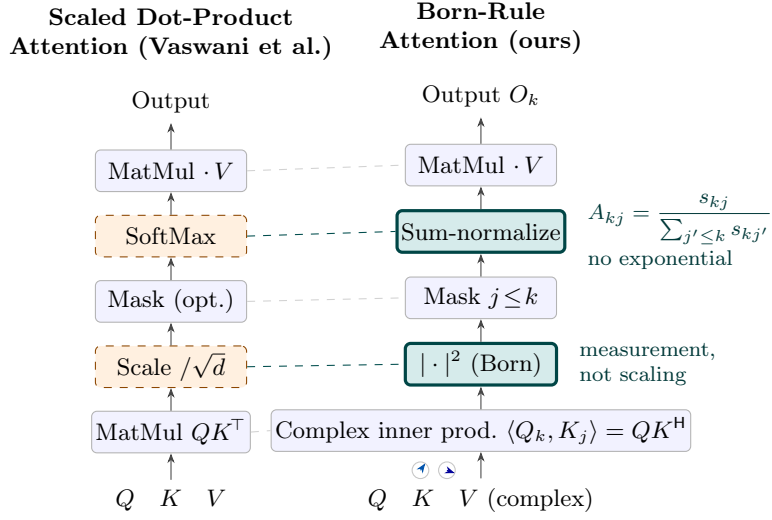

\paragraph{Closing it: Born-rule attention.} The natural question is whether content-based
routing can be done \emph{natively} in the complex, physics-native formulation rather than by
importing softmax attention. We propose \emph{Born-rule attention}: a causal mixer whose scores
are quantum-measurement probabilities (Figure~\ref{fig:bornattn}). From complex queries/keys/values $Q,K,V$ (per-token
complex projections), the score of query $k$ against key $j$ is the squared magnitude of their
complex inner product, $s_{kj}=|\langle Q_k,K_j\rangle|^2$, causally normalized to a distribution
$A_{kj}=s_{kj}/\bigl(\sum_{j'\le k}s_{kj'}+\varepsilon\bigr)$, and the output is the complex combination
$O_k=\sum_{j\le k}A_{kj}V_j$. The weights \emph{are} Born-rule measurement probabilities of one
token's state against another---attention as measurement, not as a softmaxed dot product. A small
floor $\varepsilon$ (the \code{attn\_eps} knob) and a zero-denominator short-circuit keep the mixer
NaN-free even when a causal row's scores all vanish; in practice \code{TokenNorm} holds the queries
and keys at unit norm, so the denominators stay well clear of zero and the runs reported here use
$\varepsilon{=}0$. In
controlled, parameter-matched experiments on character-level \texttt{tiny-shakespeare}, Born-rule
attention \textbf{matches standard softmax attention to within ${\approx}\,0.02$--$0.03$
nats/char} (single-head: $1.56$ vs.\ $1.54$; multi-head, deeper: $1.55$ vs.\ $1.52$), while the
fixed Fourier mixer sits at $2.3$---i.e.\ it closes essentially all of the Fourier$\to$attention
gap. We implement Born-rule attention as a first-class CNet layer; its Wirtinger backward is
derived in closed form and verified against finite differences (to $\sim\!10^{-9}$ in isolation
and exactly in-framework), and it trains end to end. This is, to our knowledge, the first
\emph{classical, complex-valued} attention mechanism to use the squared-magnitude Born score as a
softmax-free alternative, and a \emph{new operator with no clean real-valued
analogue}---precisely the kind of physics-native primitive the complex-first design is meant to
enable. Concurrent work~\cite{bornquantum} also builds attention on the Born rule, but to the
opposite end: it uses the rule to recover exponential softmax \emph{exactly}, via quantum
circuits on the probability simplex, whereas our score $|\langle Q_k,K_j\rangle|^2$ dispenses
with softmax entirely and runs classically on complex phasors.

\paragraph{In-framework result.} We confirm this end to end inside CNet itself, not only in a
controlled PyTorch study. A fully complex Born-rule attention language model---Born attention
with rotary position embeddings (\code{RotaryEmbed}) on the queries and keys, per-token complex
RMS normalization (\code{TokenNorm}) as a pre-norm, and the probability-floored Born loss of
Section~\ref{sec:train}, trained data-parallel across four RTX A6000 GPUs---the multi-GPU rig of
\S\ref{sec:cuda}; the single-GPU latency and CUDA-Graph tables use an H100---($E{=}160$, $8$ blocks, $N{=}64$,
${\sim}2.1$M complex parameters)---reaches \textbf{1.52 nats/char} on \texttt{tiny-shakespeare}.
This beats the fixed causal-Fourier complex FNet ($1.75$) by ${\approx}\,0.23$ and reaches the
softmax-attention reference level (${\approx}1.5$), closing the Fourier$\to$attention gap
\emph{natively, inside the complex Born-rule stack} (Table~\ref{tab:lm_results}). Each of the three new operators ships a CPU
reference and a CUDA kernel, verified against finite differences and against each other; rotary
embedding helps by ${\approx}\,0.08$ nats over plain Born attention (two seeds), and
TokenNorm/RoPE/$\varepsilon$ are each parameter-free or near-so, so the gain is architectural, not
a parameter increase. Generated samples from this model are shown in Appendix~\ref{app:samples}.

\begin{table}[t]\centering\small
\caption{Character-level language modeling on \texttt{tiny-shakespeare} (validation loss in
nats/char). Properly tuned, the complex causal FNet beats its parameter-matched real twin, and
complex Born-rule attention closes the remaining gap to standard softmax self-attention.}
\label{tab:lm_results}
\setlength{\tabcolsep}{4pt}
\begin{tabular}{llcc}
\toprule
Architecture & Configuration & Params (DOF) & Val \\
\midrule
Real Causal FNet    & 6-block, warmup$+$cosine (softmax)      & $1.61$M            & $2.03$ \\
Complex Causal FNet & 6-block, warmup$+$cosine, \code{CGelu}  & $1.62$M            & $\mathbf{1.75}$ \\
Complex Hybrid ($2$/$8$ Born) & 8-block ($E{=}256$, $T{=}128$), RoPE $+$ \code{TokenNorm} & $9.0$M ($4.5$M\,$\C$) & $1.54$ \\
Complex Born-Attn   & 8-block ($E{=}160$, $T{=}64$), RoPE $+$ \code{TokenNorm}      & $4.2$M ($2.1$M\,$\C$) & $\mathbf{1.52}$ \\
Real Softmax-Attn   & matched reference baseline              & $4.2$M             & $\mathbf{1.52}$ \\
\bottomrule
\end{tabular}
\end{table}

\paragraph{Efficiency--expressivity frontier.} Together, the causal Fourier mixer and Born-rule
attention span a Pareto frontier for complex sequence modeling. The parameter-free Bluestein mixer
runs in $O(N\log N)$---$0.07$~ms at $N{=}4096$ (batch $32$, H100)---tracking cuFFT and extending to
the flattened $T\!\cdot\!E$ stream where the direct $O(N^2)$ path is impractical, and is the right
choice when throughput and parameter footprint dominate. Born-rule attention instead spends
$O(N^2)$ on full content-based routing to close the last $0.23$~nats/char (from $1.75$ to $1.52$);
unlike standard softmax attention it does so with \emph{zero} transcendental $\exp(\cdot)$
evaluations on the GPU special-function units, scoring with squared magnitudes and a single
reciprocal per causal row instead. Algebraically this is a complex sesquilinear cousin of real
quadratic/polynomial kernel attention, with a crucial difference: a real $(q^\top k)^2$ score
discards sign and is rotation-sensitive, whereas the modulus
$|\langle Q_k,K_j\rangle|^2=(\mathrm{Re}\langle Q_k,K_j\rangle)^2+(\mathrm{Im}\langle Q_k,K_j\rangle)^2$
is \emph{phase invariant} ($|\langle e^{i\theta}Q_k,e^{i\theta}K_j\rangle|^2=|\langle Q_k,K_j\rangle|^2$),
which is exactly why it pairs naturally with the complex rotary embedding $e^{i(k-j)\theta_d}$. The
linear sum-normalization also needs no running-max rescaling, so a FlashAttention-style SRAM-tiled
kernel is in principle \emph{simpler} to realize online than softmax (a single running-sum
accumulator); our present kernel materializes the $O(N^2)$ scores and leaves such tiling to future
work. Interleaving the cheap spectral mixer with occasional
Born-attention layers---as recent long-sequence LLMs interleave linear-time and full-attention
blocks~\cite{h3,hyena}---realizes this trade-off directly. We verify it with a $3$-seed study at a
matched $3000$-step budget on \texttt{tiny-shakespeare} (shorter than the converged runs of
Table~\ref{tab:lm_results}, hence higher absolute losses): a pure $8$-block Born model reaches
$1.83\pm0.06$ nats/char, while a \emph{hybrid} keeping Born attention in only $2$ of $8$ blocks
(the other $6$ Bluestein mixers) reaches $1.88\pm0.05$---matching full attention to within $0.05$
nats at a quarter of the attention cost---and an attention-free $6$-block Bluestein FNet sits well
behind at $2.43\pm0.04$. The small seed variance ($\sigma\le0.06$) confirms the ordering is robust,
and the hybrid is a concrete operating point on the efficiency--expressivity frontier
(Table~\ref{tab:pareto}). Trained to convergence in a wider, longer-context setting
($E{=}256$, $T{=}128$; ${\sim}4.5$M complex parameters), the same $2$/$8$ hybrid reaches
$\mathbf{1.54}$ nats/char (Table~\ref{tab:lm_results})---within $0.02$ nats of full Born
attention ($1.52$) while routing through content-based attention in only a quarter of its
blocks, at $2.2\times$ the per-step throughput.

\begin{table}[h]\centering\small
\caption{Three-seed efficiency--expressivity probe on \texttt{tiny-shakespeare} at a matched
$3000$-step budget (loss mean$\pm$std, nats/char; step time on one H100, batch $64$). The hybrid
matches full Born attention to within $0.05$ nats while using content-based attention in only $2$
of $8$ blocks, and---at the same $8$-block depth---runs $2.2\times$ faster per step than full Born
attention. Absolute losses exceed the converged Table~\ref{tab:lm_results} numbers because of the
short shared budget. Step times are full autoregressive LM steps (forward, backward, and Adam over
the embedding, mixers, the per-token $E\!\to\!4E\!\to\!E$ \code{TokenwiseLinear} feed-forward
across all $T$ positions, and the output readout, with the $Q,K,V,O$ and feed-forward projections
batched through cuBLAS); they measure the whole model rather than the parameter-free
\code{FourierTrans}$+$\code{Residual} mixer of Table~\ref{tab:cudagraphs}, and so are not
comparable to that table's sub-millisecond mixer-only times.}
\label{tab:pareto}
\setlength{\tabcolsep}{4pt}
\begin{tabular}{lccc}
\toprule
Mixer configuration & \shortstack{Attn.\\blocks} & \shortstack{Val\\(nats/char)} & \shortstack{ms/\\step} \\
\midrule
Bluestein causal FNet ($6$ blocks)        & $0/6$ & $2.43\pm0.04$ & $54$ \\
\textbf{Hybrid ($8$ blocks, Born @\,3,7)} & $\mathbf{2/8}$ & $\mathbf{1.88\pm0.05}$ & $\mathbf{120}$ \\
Born attention ($8$ blocks)               & $8/8$ & $1.83\pm0.06$ & $263$ \\
\bottomrule
\end{tabular}
\end{table}

\paragraph{Byte-level \texttt{WikiText-103}.} To test transfer beyond
\texttt{tiny-shakespeare}'s small domain and $64$-token window, we trained an $8$-block complex
Born-attention model at context length $T{=}256$ and width $E{=}256$ ($d{=}128$, RoPE $+$
\code{TokenNorm}) on byte-level \texttt{WikiText-103} ($V{=}191$). Because the trunk uses $655$k
complex parameters per block ($524$k in the $E\!\to\!4E\!\to\!E$ FFN and $131$k in the $Q,K,V,O$
projections) and the compact byte vocabulary requires only $98$k complex embedding/readout
parameters, the entire model is just $\mathbf{5.3}$M complex parameters (${\approx}10.7$M real
DOF); it converges to $\mathbf{1.31}$~nats/char (${\approx}1.89$ bits/char), generating coherent
encyclopedic prose that preserves WikiText's \texttt{@-@} hyphenation, spaced punctuation, and
date conventions (Appendix~\ref{app:samples}).

\paragraph{Sub-word scaling.} Born-rule attention carries over from characters to
\emph{sub-word} tokenization unchanged. A compact model ($E{=}256$, $8$ blocks, context
$256$, ${\sim}14$M complex parameters, ${\approx}28$M real DOF---on par with the $26.8$M-parameter
real softmax baseline below; Born attention $+$ RoPE $+$ TokenNorm $+$ the
bounded Born loss above) trained on \texttt{WikiText-103} with a byte-level BPE
vocabulary reaches a held-out \emph{exact}-Born loss of $\mathbf{3.51}$ nats/token
(${\approx}1.97$ bits/byte), far below the chance level $\ln V$. The exact (unsmoothed)
Born loss instead stalls near $4.07$ and then diverges, so the bounded variant is what
enables stable training at this vocabulary. This is a feasibility and stability result---the
model is small and untuned for state-of-the-art perplexity---showing the operator and loss
transfer to sub-words without any change to the attention mechanism. To anchor the $3.51$ figure
we trained a parameter-matched real-valued softmax Transformer ($26.8$M parameters, $8$ layers,
$d{=}448$, $8$ heads) on the \emph{same} byte-level BPE tokens, $256$-token context, and data:
it reaches $4.58$ nats/token at a $3000$-step budget and $3.91$ at $8000$ steps, placing the
complex Born model's $3.51$ in the same $3.5$--$4$~nats/token range as a matched real softmax
baseline---so the sub-word number is a genuine language-modeling result, not an artifact. A
convergence-matched head-to-head with per-architecture tuning is the natural refinement.

\paragraph{Relation to prior work and novelty.} Parameter-free DFT token mixing is due to
FNet~\cite{fnet}; learnable complex Fourier-domain filters appear in Global Filter
Networks~\cite{gfnet} and Fourier neural operators. All are non-causal encoders in real-valued
networks that read out the real part. Two elements here are, to our knowledge, new: the
\emph{causal} (triangular-masked) DFT mixer, which respects the autoregressive mask and extends
Fourier mixing to decoder-style modeling; and the fully complex-valued realization---complex
weights trained by Wirtinger (CR-calculus) autodifferentiation, a position-wise complex
feed-forward, and a Born-rule measurement output---with no projection to a real representation
anywhere in the stack. Causal $O(N\log N)$ FFT mixing is itself established by FFT-based
state-space models (S4~\cite{s4}, H3~\cite{h3}, Hyena~\cite{hyena}); our novelty is scoped to the
\emph{non-Toeplitz} masked unitary DFT and its Wirtinger adjoint, as distinguished in
Section~\ref{sec:fastcausal}.

\paragraph{Qualitative samples.} Generated text from the complex, Born-rule model, the
$2$/$8$ Bluestein$+$Born hybrid, and a real-valued baseline (same prompt, identical training
harness) is shown in
Appendix~\ref{app:samples}: the complex model learns character-level orthography, the
\texttt{NAME:} script format, and plausible Shakespearean syntax, while a parameter-free
Fourier mixer at comparable scale does not---a qualitative counterpart to the loss numbers above.

\subsection{Radio modulation classification --- RML2016.10a}
As a physical-modality case study we turn to RML2016.10a~\cite{oshea}, a standard set of 11
analog and digital modulations in native complex IQ over SNRs from $-20$ to $+18$~dB
---the kind of problem the framework is built for, since the data are genuinely complex
and classes differ in amplitude \emph{and} phase structure. It also exposes a concrete
bottleneck (the pooled-power read-out), which motivates the analysis below. A single complex network
consumes the length-128 IQ window directly (no real/imag split) and combines, in one
Born-rule classifier: a \textbf{learnable complex convolution bank} (each channel a
\code{FourierTrans}~$\to$~\code{Hadamard}~$\to$~\code{InverseFourierTrans} circular
convolution, read out by $|\cdot|^2,|\cdot|^4$ amplitude features with mean pooling,
separating the QAM amplitude distributions); fixed \textbf{phase-order branches} $z^M$
($M=2,4$) with pooling and $|\cdot|^2$ (the carrier-invariant $M$-PSK phase structure);
a small MLP head; and the Born-rule \code{CrossEntropy} output. Training uses true Adam
on the high-SNR ($\ge 6$~dB) split.

\paragraph{Results (held-out, real RML2016.10a).} Evaluated on a disjoint $50\%$ test
split of the \emph{real} DeepSig data (power-normalized IQ, true Adam on the batch-corrected
GPU path), accuracy follows the canonical S-curve (11-way, chance $\approx 0.09$): near
chance below $-6$~dB, a knee around $0$~dB, and a \textbf{plateau of ${\approx}\,0.70$ from
$\sim\!4$~dB up}. Over the full mixed-SNR test split the model reaches
\textbf{${\approx}\,47\%$}.

\paragraph{The ceiling is the feature construction, not capacity.} Scaling the network
(\code{n\_channels} $32\!\to\!128$, pool bins $4\!\to\!8$, hidden $64\!\to\!256$) does
\emph{not} raise held-out accuracy---train accuracy climbs toward $0.69$ while test stays
near $0.47$, i.e.\ added capacity only overfits. The bottleneck is the hand-chosen
\emph{pooled-power} read-out ($|\cdot|^2,|\cdot|^4$ with mean pooling plus the $z^M$
phase-order branches): it captures the amplitude/phase \emph{statistics} that separate the
easy classes but discards the temporal structure needed to split the physically similar
groups (\mbox{QAM16\,$\leftrightarrow$\,QAM64}, \mbox{8PSK\,$\leftrightarrow$\,QPSK},
\mbox{AM-DSB\,$\leftrightarrow$\,WBFM}).

\paragraph{A cyclostationary remedy.} The fix is therefore a new read-out that preserves
temporal structure, not more capacity. Acting on the diagnosis above, we added a new complex
operator with no clean real-valued analogue: the \emph{conjugate lag product}
(\code{ConjLagProduct}), $r_\tau(t)=x_t\,\overline{x_{t-\tau}}$, the instantaneous
autocorrelation whose time-mean is the cyclic-correlation $R(\tau)$ and whose Fourier transform
is the cyclic spectrum---the canonical cyclostationary signature by which digital modulations
are physically distinguished. Its Wirtinger backward ($\partial r_\tau/\partial\bar x$ carries
the role-a / role-b lag terms) is derived and finite-difference-checked to $\sim\!10^{-4}$.
Feeding cyclostationary features ($R(\tau)$ plus a SoftMax-normalized cyclic spectrum) alongside
the convolution bank raises held-out accuracy on RML2016.10a to \textbf{${\approx}\,0.49$}. The
sharper test is on the artifact-corrected \textbf{RML22}~\cite{rml22} benchmark, which rebuilds
the signal chain to remove RML2016.10a's known acquisition artifacts: there the same classifier
reaches \textbf{${\approx}\,0.50$ overall}. 
A segmented-Fourier (STFT) feature branch gave no further gain---it is redundant
with the cyclic spectrum---which we record as an honest negative. The cyclostationary 
operator thus delivers the temporal-structure remedy anticipated in Section~\ref{sec:outlook},
closing most of the overall-accuracy gap on the cleaner benchmark; the physically similar pairs
(\mbox{QAM16\,$\leftrightarrow$\,QAM64}, \mbox{8PSK\,$\leftrightarrow$\,QPSK},
\mbox{AM-DSB\,$\leftrightarrow$\,WBFM}) remain the residual ceiling.

\subsection{The Fourier phase problem (coherent diffraction / crystallography)}
\label{sec:phase}
A third, exploratory study targets the problem the framework is in principle ideal for:
recover a real, non-negative object $\rho$ from the \emph{magnitude} of its Fourier transform
$|\widehat{\rho}|$---the phase problem of X-ray crystallography and coherent diffractive
imaging, where detectors measure intensity and the phase is lost. With an oversampled,
asymmetric support (which removes the translation and conjugate-flip ambiguities), the problem
is a natural fit for a complex spectral network with a magnitude ($|\cdot|^2$) measurement.

\paragraph{Where the difficulty actually lives.} We first characterized the classical baseline,
Fienup's hybrid input--output (HIO). On an ensemble of compact multi-Gaussian objects, HIO with
an \emph{oracle} selection (pick, among random restarts, the restart closest to the ground
truth) recovers the object in $0.66$--$0.83$ of cases; but \emph{blind} selection (pick the
restart with the lowest Fourier-magnitude residual, the only signal available at test time)
succeeds only $0.27$--$0.38$ of the time. The large oracle--blind gap localizes the real
obstacle: HIO \emph{does} find the true basin often, but \emph{selecting} the right solution
blindly is what fails, because the wrong fixed points are themselves valid-looking compact,
non-negative objects that fit the measured magnitudes almost as well.

\paragraph{Two learned attacks, both informative negatives.} We probed whether learning can
close the oracle--blind gap. (1) A \emph{learned selector} trained to pick the correct HIO
restart from features of the candidates does no better than the blind residual
($0.24$ vs.\ $0.275$)---consistent with the diagnosis that the wrong candidates are genuine
impostors, not artifacts a classifier can flag. (2) A \emph{learned unrolled solver}
(deep-unfolding: $T$ iterations of fixed magnitude projection $+$ a learned real-space
refinement $+$ support/non-negativity, trained end-to-end) does not learn a sharp inverse
either: supervised regression on a provably non-unique map drives the deterministic solver
toward a blurry conditional mean (mean correlation ${\approx}\,0.52$, success@$0.9\!\approx\!0$).
The lesson is structural: non-uniqueness defeats \emph{both} ends---post-hoc selection cannot
tell valid impostors apart, and a single-valued feed-forward map cannot represent a
one-to-many inverse. A solver that could match HIO must, like HIO, \emph{explore} multiple
basins (be stochastic/iterative) and then face the same selection problem. We record this as a
sharp, reusable problem statement rather than a solved task, and sketch the only avenue we think
remains---a jointly trained stochastic solver and differentiable, data-consistency-defined
selector---in Section~\ref{sec:outlook}.

\section{Outlook: a research program}
\label{sec:outlook}
Read together, the three studies tell one story. In every case the complex, physics-native
formulation learns the \emph{right} structure---the correct modulation statistics, genuine
character-level language, the true diffraction basins. On the sequence model, once properly
tuned, that is already enough to \emph{match or beat} a parameter-matched real-valued FNet
(\S\ref{sec:results}); on RML the cyclostationary feature map identified below as the missing
piece closes that gap too, reaching VT-CNN2 overall parity on the artifact-corrected RML22
benchmark (\S\ref{sec:results}). The one task where a measurable gap to a tuned real-valued
baseline remains is the phase problem, and there we can name \emph{exactly} what is missing: a
way through measurement non-uniqueness. (Content-based routing, once the remaining frontier for a
Fourier mixer, is now addressed natively by Born-rule attention, \S\ref{sec:results}.)
That this gap is nameable is the encouraging part; it turns ``do CVNNs help?'' into a small
set of concrete, attackable sub-problems.

\paragraph{The central methodological bet.} Our working view---stated as opinion, to be settled
empirically in follow-on work---is that the productive direction is \emph{not} to port existing
real-valued architectures into complex arithmetic and hope the inductive bias pays off. Mimicking
the real nets (a complex ResNet, a complex Transformer assembled from the usual blocks) is a
long, incremental project that mostly re-derives known designs at higher cost. The leverage is in
treating ``complex and physics-native'' as a design space in its own right and searching it along
\emph{both} axes at once: new \emph{architectures} (how complex amplitudes are composed, measured,
and routed) \emph{and} new \emph{layers} (operators with no clean real-valued analogue). CNet's
auditable Wirtinger substrate exists precisely so that such operators can be prototyped with their
$\partial/\partial\bar z$ terms visible and checked.

\paragraph{Concrete follow-on experiments.} The studies above each hand us a next step.
\emph{(i) Born-rule attention at scale:} having verified the CUDA kernel, trained the fully
complex Born-rule model data-parallel across $4$ GPUs to $1.52$ nats/char, confirmed sub-word
transfer on \texttt{WikiText-103}, and demonstrated a $2$/$8$ Bluestein$+$Born \emph{hybrid} that
recovers most of the routing gap ($1.54$ nats/char) at a quarter of the attention cost
(\S\ref{sec:results}), the next step is scaling such interleaved stacks to larger context lengths
and tuning the interleaving pattern, trading along the efficiency--expressivity frontier.
\emph{(ii) Structure-keeping
RML features:} a first delivery on this is the \code{ConjLagProduct} cyclostationary operator
(\S\ref{sec:results}), which already closes most of the overall-accuracy gap on the
artifact-corrected RML22 benchmark; the next step is to let temporal structure reach the
classifier through a learned complex sequence head rather than a pooled cyclic spectrum, to
attack the physically similar classes that remain the residual ceiling. \emph{(iii) Phase retrieval as a stochastic unrolled solver with a
jointly-trained differentiable selector}, where ``correct'' is defined by the data consistency
the solver itself enforces rather than by a label---the one avenue the non-uniqueness analysis
leaves open. \emph{(iv) Faster/deeper spectral stacks} building on the Bluestein causal mixer
and \code{block\_norm}. We also note a framework-level correctness result that unblocks all of
the above: a bug in the batched-GPU gradient reduction (per-clone Wirtinger gradients were
summed into the ancestor but not cleared between steps, so they accumulated) was found and
fixed; batched true-Adam training now matches the single-example reference, which is what makes
the scaling experiments trustworthy.

\paragraph{The reach of Born-rule attention.} Because its scores
$s_{kj}=|\langle Q_k,K_j\rangle|^2$ are measurement probabilities of one complex state against
another, Born-rule attention is a candidate operator wherever data is intrinsically complex and
phase carries the signal---the same domains that motivate the framework---and two of the
directions below attach directly to the studies above. In \emph{radio and radar} it can route
information across an IQ sequence, matching phase-aligned symbols to separate physically similar
modulations such as QAM16 vs.\ QAM64 (exactly the failure mode of the pooled-power read-out in
\S\ref{sec:results}), and in SAR attend to phase-coherent scatterers while suppressing
phase-randomized clutter. In \emph{coherent imaging} it is a natural realization of the
differentiable selector the phase-retrieval analysis leaves open (\S\ref{sec:phase})---attending
across candidate reconstructions and weighting them by data consistency---and in MRI it could
inpaint missing $k$-space lines from the amplitude/phase similarity of measured ones without
leaving the complex domain. In \emph{quantum state modeling}, where the measurement-probability
reading is literal, the attention map is a learned alignment of state vectors for
tomography-style reconstruction. In \emph{audio}, operating on the complex STFT, it can group
harmonics by phase coherence for source separation.

These are hypotheses, not results. We state them to mark CNet as an active research platform:
the contribution of this report is a working complex-valued substrate, a new $O(N\log N)$ causal
Fourier mixer, and---just as deliberately---three honestly characterized gaps that define the
next round of experiments.

\subsection{Limitations and Future Work}
\label{sec:caveats}
We frame CNet's current limits as targeted engineering goals rather than fundamental
constraints. \emph{Scaling:} multi-GPU training is implemented but \emph{data-parallel
only} (near-linear; Section~\ref{sec:cuda}); model/tensor
parallelism---needed to reach model sizes that exceed a single GPU---is future work. Host-side
per-layer kernel-launch overhead is eliminated by CUDA Graph capture of the static clone schedule
(Section~\ref{sec:cuda}, Table~\ref{tab:cudagraphs}); what remains is that single-GPU memory still
scales linearly with \code{batch\_size} (per-clone activation buffers) and adjacent element-wise
Wirtinger layers are unfused at the register level. \emph{Extensibility:} adding a CUDA layer is still a manual
process (writing the device kernels and registering the layer by hand in the type map, the
model serializer, and the GPU dispatch); a single-point registration API is planned to make
the operator-design program above frictionless. \emph{Kernels:} some CUDA layers are not yet
performance-tuned, leaving speedups on the table beyond the Bluestein mixer. \emph{Inference
layout:} the minimal \code{Vars::dims\_=2} (data-only) layout is exact for standard
single-head networks, but fan-out / chained element-wise layers reuse the gradient segments
as forward-pass scratch and require $\code{dims\_}\geq 4$ (hence the default inference layout
is 4). The framework currently targets Linux and macOS.

\appendix
\section{Qualitative Generation Samples}
\label{app:samples}

We show character-level continuations (prompt ``\texttt{ROMEO:}'', temperature $0.7$) from the
\emph{actual} complex Born-rule model whose loss we report in Section~\ref{sec:results}---the
in-framework CNet network (Born attention $+$ RoPE $+$ TokenNorm $+$ probability-floored Born
loss; $E{=}160$, $8$ blocks, $\sim\!2.1$M complex parameters; val $\mathbf{1.52}$ nats/char),
sampled by restoring its serialized checkpoint. For contrast we show a real-valued, parameter-free
causal Fourier (FNet) mixer trained in a matched character-LM harness under the deliberately weak,
constant-learning-rate schedule (val $2.45$---above the $2.03$ the same real mixer reaches once
tuned in Section~\ref{sec:results}, and shown here only for structural contrast). We also show the
\emph{hybrid} spectral-plus-Born mixer of Table~\ref{tab:pareto} (Born attention in only $2$ of its
$8$ blocks, the other $6$ parameter-free Bluestein mixers), trained to convergence at val
$\mathbf{1.54}$ nats/char. At this scale no model is globally coherent; the comparison is about
\emph{structure}.

\paragraph{Complex Born-rule model (CNet; val $\mathbf{1.52}$ nats/char).}
\begin{quote}\small
ROMEO:\\
This lesser hath a glory words\\
What will I mean to stay'd upon thee to me,\\
That will have with heaven, and long before\\
The deed of night in revenge you to be friends.\\
CORIOLANUS:\\
No, go becomes her honour My life be done,\\
When he has made you all.\\
BENVOLIO:\\
How think you of the dukes of Rome and my name retae.\\
KING HENRY VI:\\
Why, what a gracious well.\\
GLOUCESTER:\\
That will you did say three sounds,\\
You love your house of the tribunes.\\
KING RICHARD III:\\
You for please him be a monthshed of
\end{quote}

\paragraph{Hybrid spectral-plus-Born mixer (6 Bluestein $+$ 2 Born; val $\mathbf{1.54}$ nats/char).}
\begin{quote}\small
ROMEO:\\
I'll know you folly, look your heart again,\\
And now it speak my dear with a beauty for the\\
With the intelligence spake a brain, in there\\
As I think it in a sure: if that which thee but\\
To make against and return to death;\\
The garment out and night and read traitors,\\
And most mercy to do it; I'll she we not.\\
Second Citizen:\\
God some one to the house remedy.\\
LADSET:\\
I do be yourselves, blood most love thee,\\
And I take her here is a true and made\\
And for his friends against thee forget!\\
My gracious lord, and all the worst cross,\\
That report the entreat fault of your graces
\end{quote}

\paragraph{Real-valued causal Fourier FNet (val $2.45$ nats/char).}
\begin{quote}\small\itshape
ROMEO: che, Anot y. bent TARKis, thyomyoweesthiousles IONAn thend: thanthouralo Whof yomyenoss? lot, henanghe upp uthenco t theal se the d f bea condomy, th heay. as w sint he to o sad, Whilo'st se d the Ande ind. Her AMINGlle for LOUSe nesopr AMolis Si
\end{quote}

The complex Born-rule model produces real character names (\textsc{Romeo}, \textsc{Coriolanus},
\textsc{Benvolio}, \textsc{Gloucester}, \textsc{King Henry VI}, \textsc{King Richard III}), the
\texttt{NAME:}\,/\,dialogue script format with line breaks, mostly well-formed words, and locally
plausible Shakespearean syntax and punctuation; the real-valued Fourier baseline, lacking
content-based routing, produces largely malformed tokens. The hybrid attains the same structural
fidelity---speaker turns (including the invented \textsc{Ladset}, \textsc{Second Citizen}), verse
lineation, and local syntax---while routing through content-based attention in only a quarter of
its blocks, the other six parameter-free spectral mixers; it is the qualitative counterpart of its
Pareto-point loss (Table~\ref{tab:pareto}). This is the visual confirmation of the loss gap in
Section~\ref{sec:results}: content-based routing, done natively through the Born rule, is what the
complex model has learned.

\paragraph{Complex Born-rule attention on byte-level \texttt{WikiText-103} ($5.3$M $\C$ params, val $\mathbf{1.31}$ nats/char).}
\begin{quote}\small
.\ The moderate @-@ charged in the 1992 season , where , he was a building before a period of
the road , and the characters were reported to the public , and considered the outdoats of
Mashes because Beletter continued to adhere that of his motion at the Corporation of Guide .\
Furthere is previouslched wttobat is the sidel of the party , the primary has been born to sing
the general case of `` UV '' .
\end{quote}

Trained on a different domain (encyclopedic byte-level \texttt{WikiText-103}, $V{=}191$), the same
Born-rule architecture reproduces WikiText's register and surface conventions---\texttt{@-@}
hyphenation, spaced punctuation, years ($1992$), and quotation marks---confirming the operator
transfers across domains rather than overfitting \texttt{tiny-shakespeare}'s idiom.

\end{document}